\documentclass[10pt,twocolumn,letterpaper]{article}

\usepackage{iccv}              

\definecolor{iccvblue}{rgb}{0.21,0.49,0.74}
\usepackage[pagebackref,breaklinks,colorlinks,allcolors=iccvblue]{hyperref}
\usepackage{multirow}
\usepackage[utf8]{inputenc} 
\usepackage[T1]{fontenc}    
\usepackage{hyperref}       
\usepackage{url}            
\usepackage{wrapfig,lipsum,booktabs}       
\usepackage{amsfonts}       
\usepackage{nicefrac}       
\usepackage{microtype}      
\usepackage{xcolor,colortbl}         
\usepackage{multirow}
\usepackage{graphicx}
\usepackage{pifont}

\def\paperID{4466} 
\def\confName{ICCV}
\def\confYear{2025}

\title{LLaVA-3D: A Simple yet Effective Pathway to Empowering \\ LMMs with 3D Capabilities}

\author{
    Chenming Zhu$^{1,2}$ \quad
    Tai Wang$^{2,\dagger}$ \quad
    Wenwei Zhang$^{2}$ \quad
    Jiangmiao Pang$^{2}$ \quad
    Xihui Liu$^{1,\dagger}$
    \\
    \normalsize{$^{1}$The University of Hong Kong \quad
    $^{2}$Shanghai AI Laboratory}
    \\
    \tt \normalsize\textbf{\href{https://zcmax.github.io/projects/LLaVA-3D}{https://zcmax.github.io/projects/LLaVA-3D}}
    \\
    \normalsize{ $^{\dagger}$ corresponding author}
}

\begin{document}
\twocolumn[{
    \renewcommand\twocolumn[1][]{#1}
    \maketitle
    \begin{center}
        \captionsetup{type=figure}
        \includegraphics[width=\textwidth]{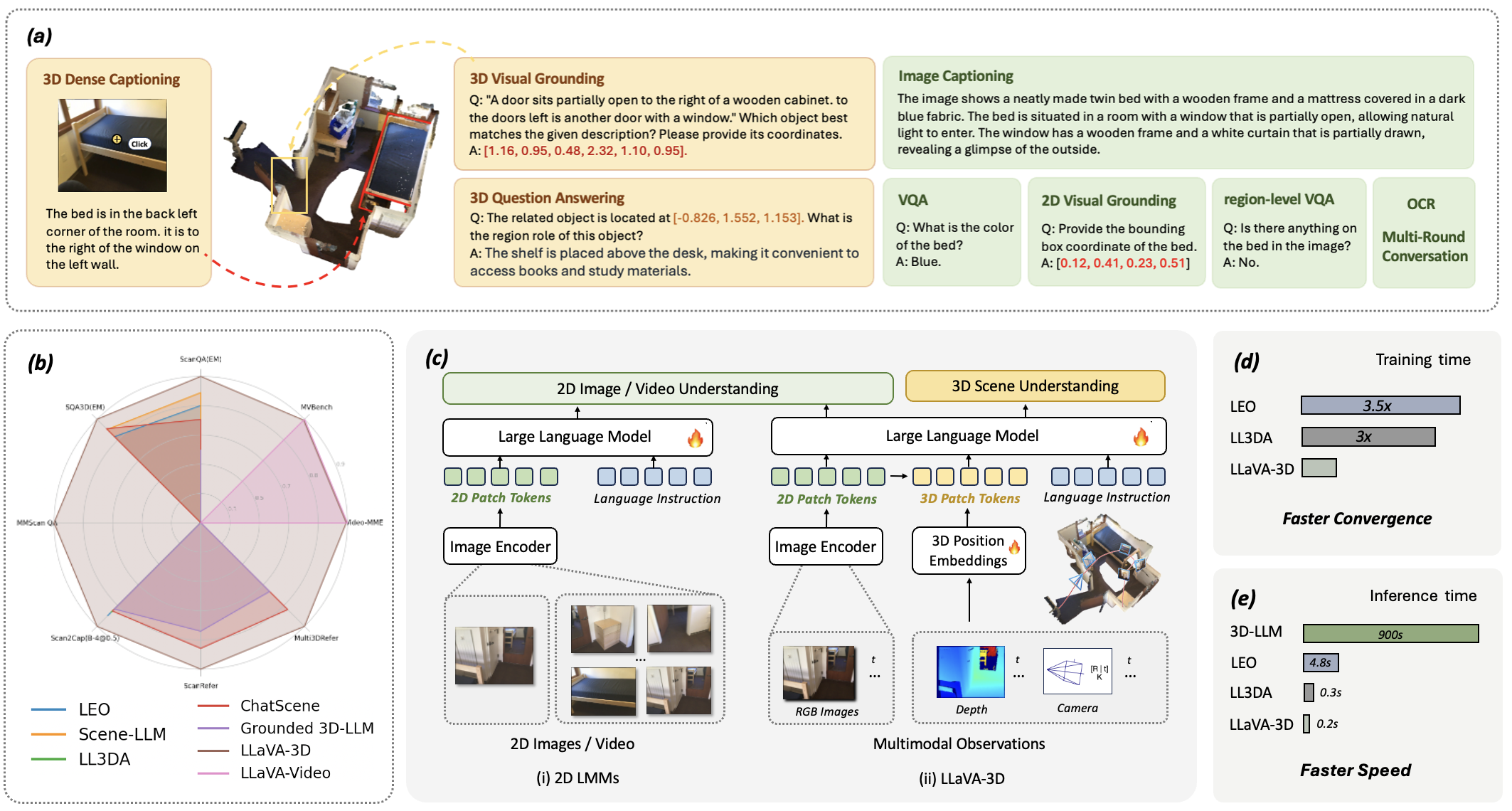}
        \caption{
            \textbf{Overview of LLaVA-3D.} The left block (b) shows that our LLaVA-3D achieves state-of-the-art performance across various 3D scene understanding benchmarks. Notably, LLaVA-3D maintains the comparable performance on 2D multimodal benchmarks compared to LLaVA-Video~\cite{llava-video}. The middle block (c) demonstrates that LLaVA-3D leverages 3D patches to endow the 2D LMMs with 3D spatial awareness, enabling it to perform various 3D vision-and-language tasks in the physical world. The right block (d) and (e) highlight the significantly faster convergence and inference speeds of LLaVA-3D compared to existing 3D LMMs.
        }
        \label{fig:new_teaser}
    \end{center}
}]

\begin{abstract}

Recent advancements in Large Multimodal Models (LMMs) have greatly enhanced their proficiency in 2D visual understanding tasks, enabling them to effectively process and understand images and videos. However, the development of LMMs with 3D scene understanding capabilities has been hindered by the lack of large-scale 3D vision-language datasets and powerful 3D encoders. In this paper, we introduce a simple yet effective framework called \textbf{LLaVA-3D}. Leveraging the strong 2D visual understanding priors from LLaVA, our LLaVA-3D efficiently adapts LLaVA for 3D scene understanding without compromising 2D understanding capabilities.
To achieve this, we utilize the 3D position embeddings to enhance the 2D CLIP Patches with 3D spatial context information and construct 3D patches. By integrating the 3D position embeddings into 2D LMMs and employing joint 2D and 3D vision-language instruction tuning, we establish a unified architecture for both 2D visual understanding and 3D scene understanding. In contrast to previous 3D LMMs, LLaVA-3D supports decoding accurate 3D spatial perception outputs, e.g., 3D bounding boxes, directly from these 3D patches, without relying on the time-consuming off-the-shelf 3D segmentors. Experimental results show that LLaVA-3D converges 3.5$\times$ faster than existing 3D LMMs when trained on 3D vision-language datasets. Moreover, LLaVA-3D not only achieves state-of-the-art performance across various 3D tasks but also maintains comparable 2D visual understanding and vision-language conversation capabilities with LLaVA.
\end{abstract}    
\section{Introduction}
\label{sec:intro}

Recent advancements in Large Multimodal Models (LMMs)~\cite{flamingo,minigpt,blip2,shridhar2023perceiver} have significantly enhanced their ability to understand and reason over visual and language inputs, leading to remarkable performance in 2D visual tasks. Despite their advanced perceptual and reasoning capabilities, LMMs are primarily confined to virtual interactions through images or video, lacking the critical ability to interact with the physical world. To enable their deployment in real-world applications and to facilitate the emergence of new capabilities through physical interactions, it is imperative to equip LMMs with 3D spatial intelligence.

A key aspect of 3D spatial intelligence is the ability to perceive and understand the 3D world. Similar to how 2D LMMs align 2D visual features with language models using large-scale 2D vision-language datasets, a common approach to developing 3D LMMs~\cite{ll3da,leo,chat3d-v2} involves integrating 3D features encoded from point clouds into Large Language Models (LLMs) and training them on 3D point cloud-language datasets. However, in contrast to the abundance of large-scale 2D datasets, 3D datasets remain relatively scarce. Meanwhile, there are no powerful pre-trained 3D point cloud encoders, akin to CLIP ViT~\cite{clip} in 2D, to provide strong and generalizable 3D features to LLMs. 

Since real-world embodied agents typically rely on ego-centric, multi-view images as raw observations, we aim to build a 3D foundation model based on such inputs rather than 3D point clouds. There have been attempts~\cite{3dllm, scene-llm} to leverage the 2D foundation models, like CLIP, alongside LLMs to advance this goal. 
These methods resort to 2D object segmentation results~\cite{sam} to extract and aggregate CLIP features from object-centric image patches, constructing pixel-aligned 3D scene features~\cite{conceptfusion}. However, this pipeline is inherently complex and computationally intensive. In contrast, 2D LMMs~\cite{minigpt,blip2,llava,llava-1.5,flamingo,llava-onevision, llava-video} directly leverage CLIP's image patch features with richer, fine-grained information for effective 2D understanding and reasoning. This naturally leads to the question: \textit{Can we directly build a 3D LMM upon the strong 2D priors from 2D LMMs, bypassing the obstacles in 3D data scale and 3D encoders}?


In light of recent progress in 2D LMMs, we propose a simple yet effective framework, \textbf{LLaVA-3D}, which extends the well-established LLaVA model to efficiently comprehend the 3D world while preserving its robust 2D multimodal perception and reasoning capabilities. Inspired by ODIN~\cite{odin}, which leverages the positional encodings for unified 2D and 3D segmentation, our LLaVA-3D encodes the 3D spatial coordinates into 3D position embeddings, and incorporate them into the 2D CLIP patches in LLaVA to construct 3D patches.
These 3D patches then undergo adaptive token compression via efficient pooling strategies before LLM processing.
To further adapt LLaVA for tasks involving accurate 3D information in input or output, we devise an effective 3D-aware position encoding and decoding approach based on the 3D patches, without relying on the off-the-shelf 3D segmentors used in previous works~\cite{chat3d-v2,chatscene,leo}.
Fine-tuned on the existing 3D vision-language datasets, our model converges rapidly and acquires 3D spatial understanding and grounding capabilities. Furthermore, the unified model architecture allows LLaVA-3D to retain the strong 2D understanding and reasoning abilities of LLaVA through joint instruction-tuning on 2D vision-language datasets. 

Although LLaVA-3D is first built upon the LLaVA-family LMMs, it is a general extension method with simple yet effective designs for equipping any 2D LMM with 3D capabilities. With preliminary attempts based on different 2D LMMs~\cite{llava-1.5,llava-video,internvl} as the foundation model, we empirically observe that LLaVA-3D can benefit from stronger perception and understanding capabilities from large-scale 2D multi-modal pretraining, especially the video-based pretraining given the intrinsic connection between video and multi-view 3D scene representations.
As illustrated in Fig.~\ref{fig:new_teaser}, experimental results demonstrate that LLaVA-3D achieves state-of-the-art performance on a wide range of 3D scene understanding benchmarks~\cite{scanqa,sqa3d,mmscan,openeqa,scan2cap,scanreason}, covering tasks such as 3D dense captioning and 3D question answering. Notably, thanks to the minimal 3D designs and powerful pretraining, LLaVA-3D attains these results with significantly less training time and fewer epochs than existing 3D LMMs, without large-scale data and pretraining costs for alignment. It could also achieve promising 3D visual grounding results without relying on the time-consuming offline 3D object preprocessing~\cite{chat3d-v2, chatscene, leo}.
Furthermore, LLaVA-3D maintains capabilities comparable to state-of-the-art 2D LMMs in 2D visual understanding, reasoning, and conversation through joint tuning on 2D and 3D vision-language instructions. 

\section{Related Work}
\label{sec:formatting}

\vspace{3pt}
\noindent\textbf{2D LMMs.} Building on the success of recent LLMs, numerous studies~\cite{llava,llava-1.5,flamingo,blip2,minigpt,vila} explored LMMs that can jointly process visual and linguistic information. For example, LLaVA~\cite{llava,llava-1.5} aligned 2D images with language models through an image encoder and a projection layer, while BLIP2~\cite{blip2} employed a sophisticated Q-Former architecture to guide the compression of visual features using textual cues. However, most early 2D LMMs were trained on single-image datasets, limiting their ability to tackle multi-image understanding. Recently, there has been increasing interest in expanding LMMs to handle multi-image inputs, addressing the demands of real-world scenarios. For video LMMs~\cite{video-llava,videochat,pllava,llava-onevision}, multi-image input forms the basis for capturing temporal or action-related dynamics across sequences of video frames. On the other hand, multi-view images of the 3D scene can implicitly reveal 3D spatial relationships and other abstract relations in the environment. Recent works ~\cite{openeqa,coarse_correspondence} explored whether 2D LMMs~\cite{gpt-4,gemini} can leverage multi-view images to perform spatial understanding. However, these methods primarily relied on implicit learning from the data, without directly modeling the 3D world. In contrast, our LLaVA-3D explicitly models the 3D world from multi-view images, enabling advanced 3D spatial understanding and grounding capabilities.

\vspace{3pt}
\noindent\textbf{Injecting 3D into LLMs.} As 2D LMMs achieved substantial progress in visual perception, similar efforts have been made in the 3D domain. For 3D scene-level understanding, recent works explored ways to integrate 3D inputs such as point clouds~\cite{chat3d-v2,ll3da,leo,chatscene} or multi-view images~\cite{3dllm,scene-llm,video3dllm,gpt4scene} into LLMs to enable advanced 3D scene understanding and reasoning. An important distinction among these methods is how they construct the 3D scene representation. LL3DA~\cite{ll3da} directly used a scene-level 3D point cloud encoder to extract the 3D scene representation. LEO~\cite{leo} and Chat-Scene~\cite{chatscene} first segmented 3D objects from the scene point cloud using the off-the-shelf 3D instance segmentation model and then independently extracted 3D object features with object-level 3D encoders to represent the 3D scene. On the other hand, starting from multi-view images, 3D-LLM~\cite{3dllm} and Scene-LLM~\cite{scene-llm} resorted to manually crafted 2D object segmentation to extract and aggregate CLIP features from object-centric image patches, constructing pixel-aligned 3D point representation. Unlike these approaches, our LLaVA-3D directly builds on the well-trained 2D LMM with multi-view images as input. Utilizing the 3D position embeddings, it brings the 2D patches within a 3D spatial context to construct 3D Patches. This 3D scene representation enables quick adaption of LLaVA for 3D scene understanding while preserving its strong 2D image understanding ability.

\vspace{3pt}
\noindent\textbf{Joint Modeling of 2D and 3D.} Recent works~\cite{petr,petrv2,odin} explored leveraging existing 2D foundation models to enhance 3D perception for detection and segmentation tasks. These methods extract 2D features from multi-view images using 2D foundation models, and then construct 3D position-aware features by incorporating 3D position embeddings for improved 3D detection and segmentation. 
ODIN~\cite{odin} utilized the posed RGB-D images as input for 3D instance segmentation. It leverages the powerful 2D pre-trained backbone and differentiates between 2D and 3D features by using distinct learnable position encodings, with 2D features represented by pixel coordinates and 3D features represented by 3D coordinates. This unified architecture facilitated joint training on both 2D and 3D datasets, further enhancing the performance of 3D segmentation. 
Our LLaVA-3D first integrates the 3D position-aware features into 2D LMMs, enabling 2D LMMs to achieve 3D understanding. Analogous to ODIN~\cite{odin}, this modeling approach enables joint 2D-3D training, allowing the model to process and reason about both 2D and 3D tasks in a unified framework.
\section{Method}





\begin{figure*}[t]
  \centering
   \includegraphics[width=\linewidth]{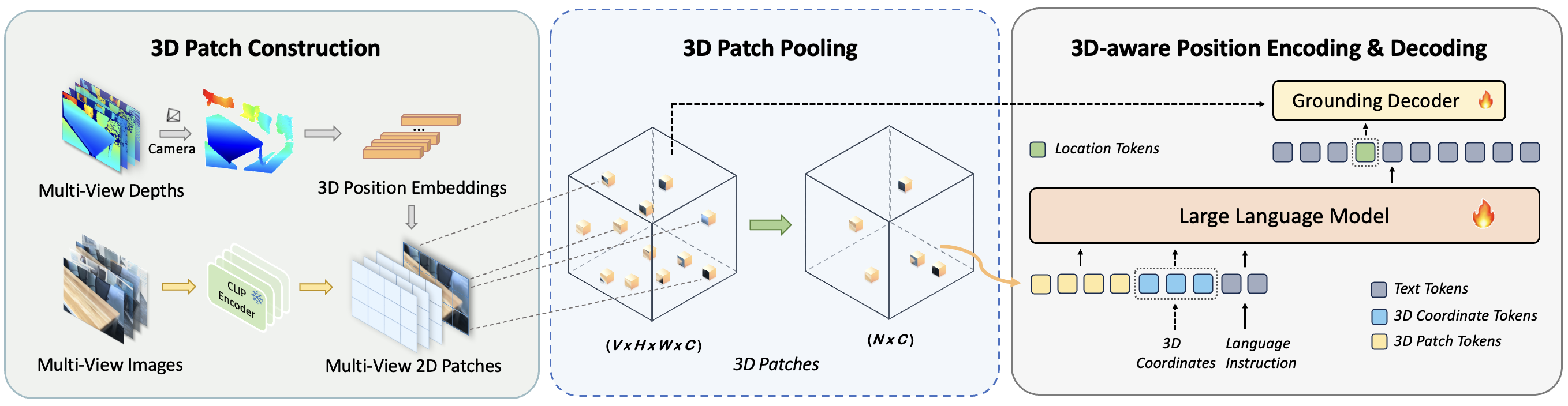}
   \captionsetup{aboveskip=10pt}\captionsetup{belowskip=0pt}\caption{\textbf{LLaVA-3D Architecture.} Based on LLaVA-Video, we directly add the corresponding 3D position embeddings to 2D patch visual tokens of multi-view images to construct the 3D patches. Considering the context length support of the base model, we skip the 3D Patch Pooling stage and directly send the 3D patches into the 3D-aware Position Encoding and Decoding process to perform various 3D understanding tasks.} 
   \label{fig:method_v2}
\end{figure*}



Previous 2D LMMs typically consist of a visual encoder to extract 2D image features, which are then aligned with the LLM via the projection layer for joint visual and language reasoning tasks. In this section, we introduce how to bridge the 2D image features within 3D spatial context to construct 3D patches (Sec.~\ref{sec:preliminary},~\ref{sec:3d_patch}), and then demonstrate the 3D-aware pooling strategies to compress the 3D patches (Sec.~\ref{sec:3d_patch}) and finally present the 3D-aware position encoding and decoding process (Sec.~\ref{sec:3d_multimodal}), as illustrated in Fig.~\ref{fig:method_v2}.

\subsection{Preliminary}
\label{sec:preliminary}
We choose LLaVA-Video~\cite{llava-video} as the base model to build the 3D LMM. For each frame image, LLaVA-Video uses the pre-trained CLIP encoder to split the image $X \in \mathbb{R}^{3 \times W \times H}$ into patches at the patch size $P$ and extract the 2D patch features $X' \in \mathbb{R}^{c \times w \times h}$, where $h = \left\lfloor \frac{H}{P} \right\rfloor, w = \left\lfloor \frac{W}{P} \right\rfloor$, and then align the 2D patch features $X_v$ into with LLM space with the projection layer. For the 3D scene understanding, the multi-view image patch features $X'_v \in \mathbb{R}^{V \times c \times w \times h}$ are sequentially sent into LLM. To empower LLaVA-Video with 3D capabilities, we incorporate the 3D position embeddings into 2D patches to obtain the 3D patches.

\subsection{3D Patch} 
\label{sec:3d_patch}
Our 3D patch representations are built upon the 2D patch features $X'_v$ extracted from multi-view images with CLIP visual encoder to leverage the strong visual-semantic alignment. To construct the 3D patches, we inject the 3D position information into the 2D patches so that the 3D patches can explicitly model 3D spatial information while preserving the semantic information from 2D patches. As illustrated in left block of Fig.~\ref{fig:method_v2}, given the multi-view 2D patch features $X'_p \in \mathbb{R}^{V \times d \times w \times h}$ after the projection layer, we obtain their 3D positions $P \in \mathbb{R}^{V \times 3 \times w \times h}$ in the 3D world, using nearest neighbor depth and known camera intrinsic and extrinsic parameters, following ODIN~\cite{odin}. The 3D positions $P$ are then encoded into 3D position embeddings $P' \in \mathbb{R}^{V \times d \times w \times h}$ through the \textit{3D Position Encoding Layer} which consists of a learnable two-layer MLP. The 3D position embeddings are subsequently added to the 2D patch visual tokens, resulting in the 3D patches $X'_{3D} \in \mathbb{R}^{V \times d \times w \times h }$:

\begin{equation}
X'_{3D} = X'_p + P'
\end{equation}



\subsection{3D Patch Pooling}
\label{sec:3d_patch_pooling}

While 3D patches enhance 2D patches with spatial information, they increase linearly with the number of input images and may exceed the context length of LLM. To address this, we introduce a 3D-aware pooling mechanism to reduce the number of 3D patches when token compression is needed, as illustrated in the middle block of Fig.~\ref{fig:method_v2}.

In the 2D image or video domain, pooling is commonly applied along the 2D spatial or temporal dimensions to compress the number of tokens and extract essential semantic information. However, for 3D scene understanding, we pool the 3D patches based on their 3D locations to ensure these features can cover and preserve the entire scene's structure as completely as possible. We explore two parameter-free pooling strategies to achieve this:

\vspace{3pt}
\noindent\textbf{Voxelization Pooling.} Voxelization discretizes the 3D space into a volumetric grid, with 3D patches undergoing average pooling within each occupied voxel, resulting in updated voxel visual tokens. Only the visual tokens from the occupied voxels are passed to the LLM, and the number of tokens varies across different 3D scenes. While the number of 3D patches scales with the number of images, the number of voxel tokens depends solely on the partitioning of the voxel grid. By adjusting the voxel size, we can effectively balance the trade-off between the number of visual tokens and the preservation of fine-grained scene features.

\vspace{3pt}
\noindent\textbf{FPS Pooling.} Farthest Point Sampling (FPS) is a widely used sampling strategy~\cite{butd-detr,groupfree3d} to select a representative subset of points from a larger set of points cloud. We apply FPS to sample 3D patches from multi-view images to a fixed number of tokens, ensuring that the sampled tokens represent the entire scene structure. While fixing the number of tokens helps the LLM efficiently process visual information, it may also result in loss of scene information.

\subsection{3D-aware Position Encoding \& Decoding}
\label{sec:3d_multimodal}

In the previous sections, we detailed the construction of the 3D scene representation from multi-view images, establishing the foundation for further interaction with the 3D scene. Building on this, the LLM could process multi-modal inputs such as the 3D scene, language instructions, and 3D coordinate cues to generate outputs such as language responses and 3D bounding boxes, as illustrated in the right block of Fig.~\ref{fig:method_v2}. In this section, we introduce how the model is equipped to interpret accurate 3D coordinate information from inputs and subsequently output precise 3D bounding boxes when specific location-related task requirements are needed.

\vspace{3pt}
\noindent\textbf{Encoding of 3D Coordinate Input.} In scenarios such as 3D dense object captioning or object-centric question answering, the language instruction contains 3D coordinates. To handle such tasks, we introduce the \textit{3D Coordinate Token} to allow the model to integrate the provided coordinates as context into its reasoning processes. Specifically, we obtain the 3D coordinate token by feeding the 3D coordinates through the 3D position encoding layer. 
The 3D coordinate tokens are fed into LLM together with 3D patch tokens and text tokens, enabling 3D coordinate-aware perception and reasoning.

\vspace{3pt}
\noindent\textbf{Decoding of 3D Bounding Box Output.} The integration of the 3D coordinate token enables the model to process 3D coordinate information from instructions effectively. However, experiments reveal that directly outputting 3D object location coordinates is non-trivial for the LLM, resulting in notably poor performance in 3D visual grounding. To handle this, previous 3D LMMs~\cite{chatscene,video3dllm,chat3d-v2} tend to rely on the offline time-consuming 3D object extraction method and convert the task to a 3D object selection task, which is not suitable for real world application. In the section, we introduce the efficient \textit{Grounding Decoder}:
The process begins with a set of instance queries sampled via Farthest Point Sampling from the 3D patches. The 3D patches contain rich language-aligned semantic information but lack detailed geometric structure. To handle the grounding task, the grounding decoder guides instance queries to capture the geometry awareness from the 3D patches and aggregate information from the LLM. Specifically, in each decoder layer, we conduct cross-attention between the instance queries and the 3D patch features and then concatenate the updated queries and location token for distance-adaptive self-attention~\cite{3d-llava} to capture the relative relationship. To model the local object geometry information at different scales and reduce the computational complexity, we apply \textit{multi-scale 3D k-NN attention} with relative 3D positional embeddings when cross-attending to 3D patch features. The updated instance queries are sent to the grounding head to predict corresponding 3D bounding boxes, and the matching score is calculated based on the similarity of the queries and location token. More details are provided in our supplementary material.

\section{Training}

To achieve strong 3D understanding and grounding capabilities without compromising 2D understanding, we conduct the two-stage training strategy, inspired by ~\cite{visionllmv2}. The first stage equips the model with various 2D and 3D task instruction following capabilites, and the second stage further enhances the 3D grounding capability.

\begin{figure}[t]
  \centering
   \includegraphics[width=1\linewidth]{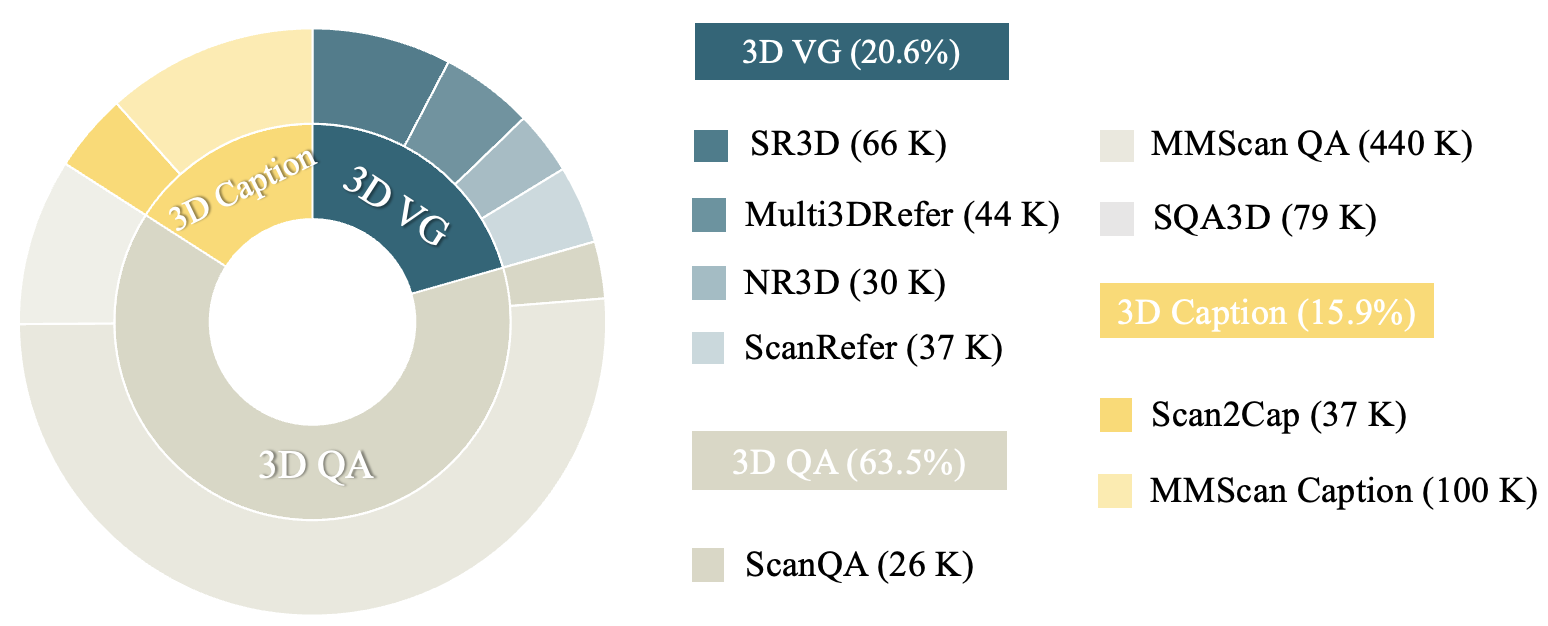}
    \captionsetup{aboveskip=10pt}\captionsetup{belowskip=0pt}\caption{\textbf{LLaVA-3D-Instruct-86K.} The 3D Dataset Collection. Left: Distribution of data across categories, with the inner circle representing all categories and the outer circle illustrating data subset distribution. Right: Detailed dataset quantities.} 
   \label{fig:3d_instruct_data}
\end{figure}

\vspace{3pt}
\noindent\textbf{Stage 1: Multi-Task Instruction Tuning.} During the instruction-tuning stage, LLaVA-3D is optimized to respond to complex 3D V\&L tasks while maintaining its original 2D reasoning and instruction-following capabilities. To facilitate this capability, we fine-tune the model on the joint 2D and 3D data. Specifically, the 2D data is sampled from the LLaVA-Video training dataset, ensuring the preservation of 2D video comprehension and conversation abilities. For the 3D data, we collect the \textbf{LLaVA-3D-Instruct-86K} dataset, a hybrid collection of 3D data specifically tailored for instruction tuning covering various 3D tasks. The overall distribution of the dataset collection is shown in Fig~\ref{fig:3d_instruct_data}. For the 3D tasks, the 3D position encoding layer will be added to jointly train with the other modules. Additionally, for tasks which requires 3D bounding box outputs, the grounding decoder will be trained together. This training setup ensures that LLaVA-3D can effectively process both 2D and 3D visual tokens and is adaptive to various tasks.



\vspace{3pt}
\noindent\textbf{Stage 2: Decoder-only Fine-tuning.} 
Since the grounding decoder fails to converge within a single epoch during the first stage of training, we further train the grounding decoder for additional epochs using the 3D visual grounding data while keeping all other components frozen. During this stage, the location token is further trained with the grounding decoder for better 3D visual grounding performance, without influencing the strong 2D and 3D scene understanding capabilities.

\section{Experiments}

\begin{table}[t]
\centering
\caption{\textbf{Quantitative comparison with SOTA models on various 3D QA tasks}. ``C'' stands for ``CIDEr'', ``B-4'' for ``BLEU-4'', ``M'' for ``METEOR'', ``R'' for ``ROUGE'', and ``EM@1'' for top-1 exact match. \textcolor{gray}{Gray} indicates evaluation results with refined exact-match protocol.}
\vspace{0.1em}
\resizebox{\linewidth}{!}{
\begin{tabular}{lcccccccccc}
    \toprule
     & \multicolumn{5}{c}{ScanQA (val)} & SQA3D (test) \\
     \cmidrule(lr){2-6} \cmidrule(lr){7-7}
      & C & B-4 & M & R & EM@1 & EM@1\\
    \midrule
    \multicolumn{1}{l}{\small\textbf{\textit{Task-specific models}}} \\
    Scan2Cap~\cite{scan2cap} & - & - & - & - & - & \hspace{2pt} 41.0$^\dagger$ \\
    ScanRefer+MCAN~\cite{scanrefer_mcan} & 55.4 & 7.9 & 11.5 & 30.0 & 18.6 & - \\
    ClipBERT~\cite{clipbert} & - & - & - & - & - & 43.3 \\
    ScanQA~\cite{scanqa} & 64.9 & 10.1 & 13.1 & 33.3 & 21.1 & 47.2 \\
    3D-VisTA~\cite{3d-vista} & 69.6 & 10.4 & 13.9 & 35.7 & 22.4 & 48.5 \\
    \midrule
    \multicolumn{1}{l}{\small\textit{\textbf{3D LMMs}}} \\
    3D-LLM (FlanT5)~\cite{3dllm} & 69.4 & 12.0 & 14.5 & 35.7 & 20.5 \\
    LL3DA~\cite{llava} & 76.8 & 13.5 & 15.9 & 37.3 & -  \\
    Chat-3D v2~\cite{chat3d-v2} & 87.6 & 14.0 & - & - & - & 54.7\\
    LEO~\cite{leo} &101.4 & 13.2 & 20.0 & 49.2 & 24.5 & 50.0 \\
    Scene-LLM ~\cite{scene-llm} & 80 & 12.0 & 16.6 & 40.0 &27.2  & 54.2\\
    ChatScene ~\cite{chatscene} & 87.7 & 14.3 & 18.0 & 41.6 & 21.6  & 54.6 \\
    
    \midrule
    \multicolumn{1}{l}{\small\textit{\textbf{Zero-shot 2D LMMs}}} \\
    VideoChat2~\cite{videochatv2} & 49.2 & 9.6 & 9.5 & 28.2 & 19.2 & 37.3 \\
    LLaVA-NeXT-Video~\cite{llava-onevision} & 46.2 & 9.8 & 9.1 & 27.8 & 18.7 & 34.2 \\
    LLaVA-Video~\cite{llava-video} & 88.7 & - & - & - & - & 48.5 \\
    GPT-4V & 59.6 & - & 13.5 & 33.4 & - & - \\
    Gemini & 68.3 & - & 11.3 & 35.4 & - & - \\
    Claude & 57.7 & - & 10.0 & 29.3 & - & - \\
    \midrule
    \rowcolor{gray!20} LLaVA-3D & \textbf{103.1} & 
    \textbf{16.4} & \textbf{20.8} & \textbf{49.6} & \textbf{30.6} & \textbf{60.1} \\
    \bottomrule
\end{tabular}
}
\label{tab:scanqa_sqa3d}
\end{table}

\begin{table*}
\centering
\caption{\textbf{Quantitative comparison on MMScan QA benchmark}. ``ST'' stands for Single-target, ``attr'' for attribute, ``OO'' for Object-Object, and ``OR'' for Object Region.``S.-BERT'', ``B-1'', ``B-4'', ``R.-L.'', ``MET.'' represents ``Sentence-BERT", ``BLEU-1'', ``BLEU-4'', ``ROUGE-L'', ``METEOR'', respectively. Here, we report the top-1 exact match with (the refined exact-match protocol results) for ``EM@1''.}
\resizebox{\textwidth}{!}
{
    \begin{tabular}{c|c|c|cc|ccc|c|>{\columncolor[HTML]{EFEFEF}}c 
>{\columncolor[HTML]{EFEFEF}}c|>{\columncolor[HTML]{EFEFEF}}c>{\columncolor[HTML]{EFEFEF}}c>{\columncolor[HTML]{EFEFEF}}c>{\columncolor[HTML]{EFEFEF}}c>{\columncolor[HTML]{EFEFEF}}c}
    \toprule
     \multirow{2}*{Methods} & \multirow{2}*{Setting} & \multirow{2}*{Overall} & \multicolumn{2}{c|}{Single-target} & \multicolumn{3}{c|}{Inter-target} &\multirow{2}*{Advanced} & \multicolumn{2}{>{\columncolor[HTML]{EFEFEF}}c|}{Data-driven Metrics} & \multicolumn{5}{>{\columncolor[HTML]{EFEFEF}}c}{Traditional Metrics}\\
    \cline{4-8}\cline{10-16}
    ~ & ~ & ~ & ST-attr & ST-space & OO-attr & OO-space & OR & ~ & SimCSE & S.-BERT & B-1. & B-4. & R.-L & MET. & EM@1 \\
    \hline
    3D-LLM~\cite{3dllm} & \multirow{4}*{Zero-Shot} & 28.6  & 37.8 & 18.8 & 13.7 & 26.3 & 15.4 & 20.8 &40.4 & 40.3 & 13.4 & 1.5 & 17.3 & 6.0 & 6.2 (19.6) \\
    Chat3D-v2~\cite{chat3d-v2} & ~ & 27.9 & 38.1 & 18.3 & 9.3 & 22.4 & 13.5 & 25.4 &45.4 & 46.3 & 18.0 & 3.0 & 22.9 & 7.5 & 10.2 (19.6) \\
    LL3DA~\cite{ll3da} & ~ & 15.8 & 15.5 & 14.7 & 14.2 & 25.2 & 4.3 & 6.4  & 40.7 & 43.6 & 5.4 & 2.1 & 16.4 & 4.4 & 8.3 (19.4) \\
    LEO~\cite{leo} & ~ & 22.2 & 28.9 & 17.6 & 18.1 & 20.4 & 15.0 & 16.3 &40.4 & 41.0 & 11.0 & 0.7 & 17.1 & 4.9 & 9.6 (18.7) \\
    \hline
    LL3DA~\cite{ll3da} & \multirow{2}*{Fine-tuning} & 38.5 & 40.4 & 46.2 & 14.7 & 47.1 & 26.4 & 7.1 & 65.3 & 67.0 & 26.4 & 8.5 & 44.3 & 14.7 & 30.2 (37.6)\\
    LEO~\cite{leo} & ~ & 47.8 & 55.5 & 49.5 & \textbf{36.1} & 45.6 & 32.1 & 38.4 & 71.2 & 72.2 & 32.0 & 12.5 &  52.1 & 17.7  & 36.6 (44.5)\\
    \hline
    LLaVA-3D & Generalist & \textbf{55.4} & \textbf{63.2} & \textbf{57.1} & 34.1 & \textbf{63.2} & \textbf{47.5} & \textbf{44.9} & \textbf{76.2} & \textbf{78.3} & \textbf{39.2} &\textbf{13.9}	&\textbf{57.5} &\textbf{20.3}	&\textbf{50.1 (54.9)} \\
    \bottomrule
    \end{tabular}
}
\vspace{-1.5ex}
\label{tab:mmscan}
\end{table*}

\begin{table}
\centering
\caption{\textbf{Quantitative comparison with SOTA models on OpenEQA benchmark.}}
\resizebox{0.55\linewidth}{!}{
\begin{tabular}{c|c|c}
      \toprule
      Models & Frame & Accuracy \\
      \midrule
      LLaMA2~\cite{touvron2023llama} & 0 & 28.3 \\
      GPT-4~\cite{gpt-4} & 0 & 33.5 \\
      Claude3 & 20 & 36.3 \\
      Gemini-Pro~\cite{gemini} & 15 & 44.9 \\
      GPT-4V~\cite{gpt-4} & 15 & 54.6 \\
      GPT-4V~\cite{gpt-4} & 50 & 55.3 \\
      Human & Full & 86.8 \\
      \midrule
      \rowcolor{gray!20} LLaVA-3D & 32 & 53.2\\
      \bottomrule
    \end{tabular}
}
\label{tab:openeqa_results}
\end{table}

In this section, we conduct extensive evaluations to examine the capabilities of LLaVA-3D, based on LLaVA-Video. We compare our model’s 3D scene understanding (Sec.~\ref{sec:qa_results},~\ref{sec:dc_results},~\ref{sec:vg_results}) and 2D video understanding (Sec.~\ref{sec:2d_benchmark}) capability with previous methods. Then we thoroughly analyze the effectiveness of the components and designs of LLaVA-3D and demonstrate the our LLaVA-3D can be integrated with any 2D LMMs to equip 3D capabilities (Sec.~\ref{sec:arch_analysis}).



\subsection{Evaluation on 3D Question Answering}
\label{sec:qa_results}

3D Question Answering requires a model to generate responses to the natural language questions in a 3D scene. In this section, we validate LLaVA-3D performance on various 3D question answering benchmarks: ScanQA~\cite{scanqa}, SQA3D~\cite{sqa3d}, MMScan QA~\cite{mmscan}, and OpenEQA~\cite{openeqa}.

\vspace{3pt}
\noindent\textbf{Spatial Understanding with ScanQA and SQA3D.} ScanQA and SQA3D are both built on the ScanNet dataset. The ScanQA dataset consists of 41363 questions about 800 scenes, including 32337 unique questions. SQA3D comprises 20.4k descriptions of 6.8k unique situations collected from 650 ScanNet scenes and 33.4k questions about these situations. Questions in ScanQA require basic recognition and 3D reasoning capabilities, and SQA3D further incorporates situation understanding and situated reasoning into embodied 3D scene understanding. 
As shown in Tab.~\ref{tab:scanqa_sqa3d}, LLaVA-Video could achieve promising performance on these benchmarks in a zero-shot manner, even surpassing the task-specific methods. This phenomenon may suggest that these benchmarks do not truly assess the model's 3D spatial understanding ability. Notably, our model could achieve the SOTA performance on these benchmarks.

\vspace{3pt}
\noindent\textbf{Coordinate Spatial Understanding with MMScan QA.} MMScan QA includes 5.2k scans from ScanNet, 3RScan, and Matterport3D, along with 116k training questions and 29k validation questions. These questions span existential inquiries, attribute understanding, and more advanced queries. Unlike ScanQA and SQA3D, some MMScan QA questions require 3D reasoning based on object coordinates rather than relying solely on text descriptions, demanding the model capable of understanding 3D coordinates information. 
We present the results under GPT-4 evaluation, data-driven metrics, and traditional metrics respectively in Tab.~\ref{tab:mmscan}. Our LLaVA-3D achieves significantly better performance compared to LL3DA and LEO which are further fine-tuned on full MMScan QA. The results highlight the training efficiency of LLaVA-3D and its strong 3D understanding ability to serve as the generalist model.

\vspace{3pt}
\noindent\textbf{Embodied Question Answering with OpenEQA.} OpenEQA is the first open-vocabulary benchmark designed for spatial understanding and embodied reasoning in embodied question answering. It features an automated evaluation protocol powered by LLMs, which shows strong alignment with human judgment. Our evaluations are conducted using the EM-EQA data split of OpenEQA, which includes over 1,600 high-quality, human-generated questions from diverse real-world environments. 
The results in Tab.~\ref{tab:openeqa_results} demonstrate that LLaVA-3D surpasses Claude3 and Gemini-Pro, and achieves comparable performance with powerful GPT-4V on this benchmark with significantly fewer model parameters.

\begin{table}[t]
    \caption{
        \textbf{Quantitative Comparisons with SOTA models for 3D Dense Captioning on Scan2Cap.} The n-gram metrics for Scan2Cap are governed by IoU@0.5.
    }
    \label{tab:benchmark-3d-dense-cap}
    \centering
    \resizebox{0.95\linewidth}{!}{
    \begin{tabular}{lcccc}
    \toprule
    & \multicolumn{4}{c}{Scan2Cap (Val)} \\ 
    \cline{2-5}  
                            & C@0.5$\uparrow$ & B-4@0.5$\uparrow$ & M@0.5$\uparrow$ & R@0.5$\uparrow$ \\ \hline
    Scan2Cap~\cite{scan2cap}                & 39.1           & 23.3             & 22.0           & 44.8      \\
    3D-VLP~\cite{3dvlp}              & 55.0           & 32.3             & 24.8           & 51.5          \\
    3D-VisTA~\cite{3d-vista}        & 61.6          & 34.1           & 26.8         & 55.0     \\
    Vote2Cap-DETR~\cite{vote2cap-detr}       & 61.8           & 34.5            & 26.2          & 54.4         \\
    LL3DA~\cite{ll3da}          & 65.2            & 36.8      & 26.0          & 55.0                  \\
    LEO~\cite{leo}    &68.4 &36.9 &27.7 &57.8 \\
    ChatScene~\cite{chatscene} &77.2 &36.3 &28.0 & 58.1 \\
    \rowcolor{gray!20} LLaVA-3D  &\textbf{84.1} & \textbf{42.6} &\textbf{29.0}  & \textbf{63.4} \\
    \bottomrule
    \end{tabular}
    }
    \label{exp:comparison_on_scanrefer}
\end{table}

\begin{table}
\centering
\caption{\textbf{Quantitative comparison with SOTA models on the MMScan Captioning benchmark.}}
\resizebox{\linewidth}{!}
{
\begin{tabular}{c|c|cccccc|c}
\toprule
\multirow{1}{*}{model} & \multicolumn{1}{c|}{Evaluator}
& \multicolumn{1}{c}{Type} & \multicolumn{1}{c}{Color} & Shape & Position & Function & Design & \multicolumn{1}{c}{Overall} \\ \hline
LL3DA~\cite{ll3da} & GPT  & 10.0 & 26.3 & 40.6  & 38.9 & 67.5 & 21.7 & 33.6 \\
LEO~\cite{leo}  & GPT  & 34.9 & 29.7 & 63.0  & 63.7 & 75.0 & 42.7 & 51.3 \\
\rowcolor{gray!20} LLaVA-3D & GPT & \textbf{39.9} & \textbf{79.2} & \textbf{89.1} & \textbf{82.2} & \textbf{94.1} & \textbf{88.0} & \textbf{78.8} \\
\bottomrule
\end{tabular}
}
\label{tab:mmscan_caption}
\end{table}

\subsection{Evaluation on 3D Dense Captioning}
\label{sec:dc_results}

3D dense captioning requires the model to localize all the objects in a 3D scene and then generate a descriptive sentence for each object. To evaluate our model on the 3D dense captioning tasks, we utilize the off-the-shelf 3D instance segmentation model~\cite{mask3d} to generate object proposals. Then we further construct the 3D coordinate tokens based on the 3D object center coordinates to guide the model to handle the task. We report the performance of various methods on two 3D dense captioning benchmarks:

\vspace{3pt}
\noindent\textbf{Scan2Cap.} Scan2Cap requires the model to describe the object's appearance and the spatial relations with nearby objects and output the corresponding 3D bounding box. As illustrated in Tab.~\ref{exp:comparison_on_scanrefer}, our method consistently outperforms the existing method on the Scan2Cap benchmark. 

\vspace{3pt}
\noindent\textbf{MMScan Captioning.} MMScan Captioning focuses on identifying common aspects of 3D objects such as Object Type, Color, Shape, Position, Function, and Design. 
We benchmark various methods on MMScan Captioning benchmark in Tab.~\ref{tab:mmscan_caption}. The results show that our method surpasses existing approaches across all metrics by a substantial margin, especially achieving a 49.5\% and 43.3\% improvement in the Color score and the Design score respectively. The strong performance further demonstrates the advantages of architectures based on 2D LMMs.

\begin{figure*}[t]
  \centering
   \includegraphics[width=1\linewidth]{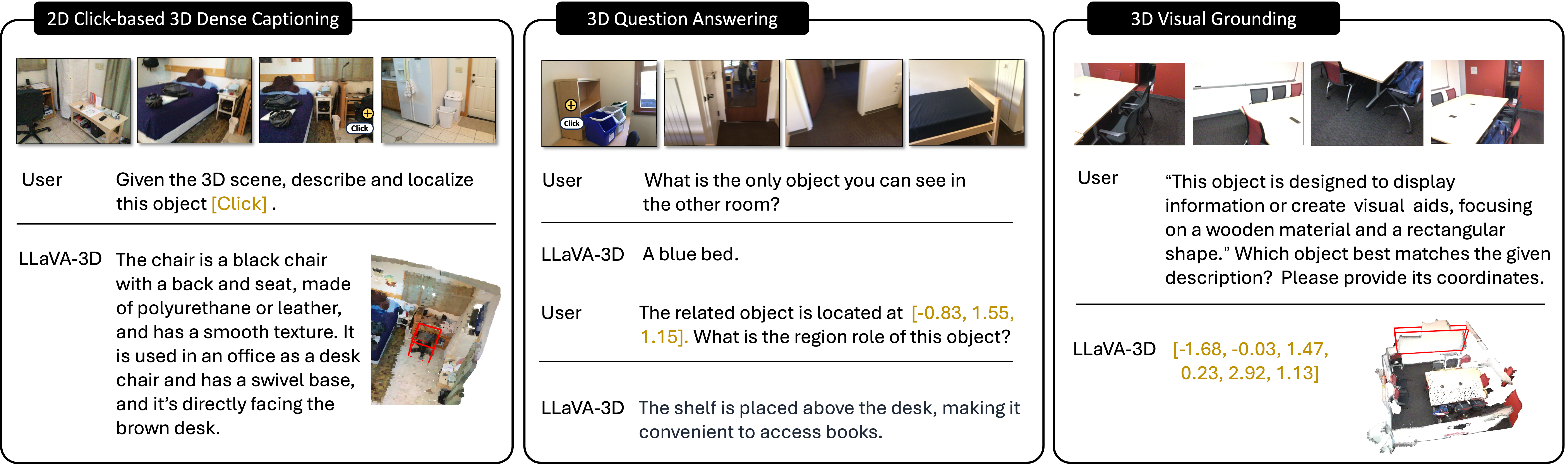}
   \captionsetup{aboveskip=5pt}\captionsetup{belowskip=0pt}\caption{LLaVA-3D enables the user-friendly interaction with the 3D scene across various 3D understanding and reasoning tasks. It allows the users to just click on the 2D images or the video frame to simply conduct the interactive 3D question answering and 3D dense captioning.} 
   \label{fig:dense_cap}
\end{figure*}

Uniquely, LLaVA-3D takes multi-view images as inputs, enabling a user-friendly feature where users can simply click on the selected images to generate both 3D object captions and 3D bounding boxes, as illustrated in Fig.~\ref{fig:dense_cap}.

\subsection{Evaluation on 3D Visual Grounding}
\label{sec:vg_results}


3D visual grounding aims to localize the target object in the 3D scene using the natural language description. In this section, we initially report the performance on the ScanRefer~\cite{scanrefer} and Multi3DRefer~\cite{zhang2023multi3drefer} benchmarks in Tab.~\ref{tab:vl_results}. Previous task-specific methods~\cite{3d-vista, butd-detr, scanreason} typically rely on point clouds extracted from 3D reconstructed meshes as input, which are not easily accessible in real-world applications. Meanwhile, current 3D LMMs~\cite{chatscene,chat3d-v2, grounded-3d-llm} tend to decouple the grounding task into two stages: first, extract the 3D objects from the 3D scene using the off-the-shelf 3D segmentor and then convert the task into a 3D object selection task. Without relying on the constructed point cloud, our method could directly decode the accurate 3D bounding boxes from 3D patches and achieve the SOTA performance (49.8 Acc@0.25 on Multi3DRefer) in the single-stage manner.

\begin{table}
\centering
\caption{\textbf{Quantitative comparison with SOTA models on various 3D VG tasks}. $\dagger$ represents that we apply the current two stage 3D visual grounding approach to LLaVA-3D.}
\vspace{0.1em}
\resizebox{0.85\linewidth}{!}{
\begin{tabular}{lcccc}
    \toprule
     & \multicolumn{2}{c}{ScanRefer} & \multicolumn{2}{c}{Multi3DRefer}\\
     \cmidrule(lr){2-3} \cmidrule(lr){4-5}
     & Acc@0.25 & Acc@0.5 & Acc@0.25 & Acc@0.5\\
    \midrule
    \multicolumn{1}{l}{\small\textbf{\textit{Task-specific models}}} \\
    ScanRefer~\cite{scanrefer}	&37.3	&24.3 & - & - 	\\
    MVT~\cite{mvt}	&40.8	&33.3 & - & - \\
    3DVG-Trans~\cite{3dvg-trans} &45.9	&34.5 & - & - \\
    ViL3DRel~\cite{vil3dref}	&47.9	&37.7 & - & - \\  
    BUTD-DETR~\cite{butd-detr}	&52.2 &39.8 & - & - \\  
    ReGround3D~\cite{scanreason}	&53.1 &41.1	& - & - \\ 
    M3DRef-CLIP~\cite{scanreason}	&51.0 &44.7	 &42.8 &38.4\\ 
    \midrule
    \multicolumn{1}{l}{\small\textit{\textbf{Two-Stage 3D LMMs}}} \\
    {\color[HTML]{969696}Chat-3D v2}~\cite{chat3d-v2}	&{\color[HTML]{969696}35.9}	&{\color[HTML]{969696}30.4}  &- &- \\
    {\color[HTML]{969696}Grounded 3D-LLM}~\cite{grounded-3d-llm}   &{\color[HTML]{969696}47.9}  &{\color[HTML]{969696}44.1} &{\color[HTML]{969696}45.2} &{\color[HTML]{969696}40.6}\\ 
    {\color[HTML]{969696}Chat-Scene}~\cite{chatscene}   &    {\color[HTML]{969696}55.5}  &    {\color[HTML]{969696}50.2} &    {\color[HTML]{969696}57.1} &    {\color[HTML]{969696}52.4} \\ 
    {\color[HTML]{969696}LLaVA-3D}$^\dagger$   &    {\color[HTML]{969696}63.9}  &    {\color[HTML]{969696}58.6} &    {\color[HTML]{969696}68.1} &    {\color[HTML]{969696}62.9} \\
    \midrule
    \multicolumn{1}{l}{\small\textit{\textbf{Single-Stage 3D LMMs}}} \\
    3D-LLM~\cite{3dllm}  &30.3	&-  &- &- \\
    \rowcolor{gray!20} LLaVA-3D	& 50.1 & 42.7 &\textbf{49.8} &\textbf{43.6}\\
    \bottomrule
\end{tabular}
}
\label{tab:vl_results}
\end{table}

\begin{table}[t!]
\centering
\caption{\textbf{Quantitative Comparisons on 2D video benchmarks}.}
\resizebox{0.6\linewidth}{!}{
\begin{tabular}{ccccccccccc}
\toprule
Method & {\bf MVBench} & {\bf VideoMME}  \\
\midrule
LLaVA-Video~\cite{llava-video}  & 58.6 & 63.3   \\
\midrule
\rowcolor{gray!20} LLaVA-3D  & 58.1 & 62.8  \\
\bottomrule
\end{tabular}
}
\label{tab:2d_benchmark}
\end{table}

\subsection{Evaluation on 2D benchmarks}
\label{sec:2d_benchmark}

Since our model is trained on the joint 2D and 3D datasets, we evaluate it on the 2D video benchmarks to ensure it retains the 2D understanding capabilities of the original LLaVA-Video. As demonstrated in Tab.~\ref{tab:2d_benchmark}, LLaVA-3D achieves performance comparable to LLaVA-Video across various 2D video understanding and conversation benchmarks, which the current existing 3D LMMs do not possess. This performance highlights the architectural advantages of our model over other 3D LMMs.

\begin{table*}[t]
    \caption{
        \textbf{Analysis of model architecture and generalization} on various 3D V\&L Benchmark.}
    \centering
    \resizebox{\linewidth}{!}{
        \begin{tabular}{cccccccc}
        \toprule
               & 3D Feature & Connector & LLM / LMM  & ScanQA   & SQA3D  & ScanRefer & Inference time \\ 
        \hline 
        (a)    & (SAM + CLIP) w / PE & Q-Former & Vicuna-7B  & 21.9 & 49.3  & - & 900s      \\
        (b)    & (SAM + CLIP) w / PE & Pooling + MLP & Vicuna-7B & 22.1 & 49.2    & - & 900s \\
        (c)    & CLIP w / PE   & Pooling + MLP & Vicuna-7B    & 23.4 & 51.2  & 43.8 & 0.2s\\
        (d)    & CLIP w / PE   & Pooling + MLP  & LLaVA-1.5-7B~\cite{llava-1.5}   & 27.0 & 55.6   & 47.9 & 0.2s \\
        (e)    & CLIP w / PE   &  MLP  & InternVL2.5-7B~\cite{internvl}       & 29.1 & 58.8 & 49.3 & 0.3s\\ 
        \rowcolor{gray!20}(f)    & CLIP w / PE   & MLP  & LLaVA-Video-7B~\cite{llava-video}       & \textbf{30.6} & \textbf{60.1} & \textbf{50.1} & 0.2s\\ 
        \bottomrule
        \end{tabular}
    }
    \label{exp:comparison_on_arch}
\end{table*}



\begin{table}[t]
    \caption{
        \textbf{Effectiveness of 3D Patch Representation}}
    \centering
    \resizebox{\linewidth}{!}{
        \begin{tabular}{cccccc}
        \toprule
        Patch Type  & ScanQA   & SQA3D  & MMScan QA & Scan2Cap \\ \hline
        2D  & 29.4 & 59.8 &  42.1   & 29.7 \\
        \rowcolor{gray!20} 3D & 29.8 (+0.4) & 60.1 (+0.3) & 55.4 (+13.3)  & 84.1 (+54.4)  \\
        \bottomrule
        \end{tabular}
    }
    \label{tab:3d_patch}
\end{table}

\subsection{More Analysis}
\label{sec:arch_analysis}

In this section, we first delve deeper into the architectural benefits and efficacy of adapting the 2D LMM to 3D, as opposed to developing a 3D LMM solely from LLMs. Then, we analyze the effectiveness of the components and validate the generalizability of our method on various 2D LMMs.


\vspace{3pt}
\noindent\textbf{Developing 3D LMM from LLM or 2D LMM.} As shown in Tab.~\ref{exp:comparison_on_arch}, (a), (b), and (c) develop 3D LMM from the LLM with different 3D scene feature and 3D-language connectors, while (d) builds 3D LMM from well-trained 2D LMM: LLaVA-1.5~\cite{llava-1.5}. Due to the context length limitation of the LLM in (b), (c), and (d), we utilize 3D voxelization pooling to compress the 3D patch tokens. The results demonstrate: 1) the Q-Former (a) and Pooling + MLP (b) share a similar performance on 3D V\&L benchmarks. 2) using CLIP (c) alone instead of SAM + CLIP (b) like 3D-LLM~\cite{3dllm} achieves better performance and significantly reduces 3D scene feature construction time from 900s to 0.2s. 3) Developing from 2D LMM (d) instead of LLM (c) could greatly improve the 3D scene understanding performance, achieving improvements
of 4.1\% Acc@0.25 on ScanRefer. 

\vspace{3pt}
\noindent\textbf{Effectiveness of 3D Patch.} To further ascertain the effectiveness of our proposed 3D Patch, we conduct additional experiments across a variety of 3D question answering and 3D dense captioning benchmarks. As shown in Tab.~\ref{tab:3d_patch}, injecting 3D position information into 2D patches only brings minor improvements on ScanQA and SQA3D benchmarks. However, 3D patches prove instrumental in tasks requiring objects position information and more difficult 3D spatial understanding, yielding substantial performance gains of 13.3\% on the MMScan QA benchmark and a remarkable 54.4\% improvement on the Scan2Cap benchmark.

\vspace{3pt}
\noindent\textbf{Effectiveness of Grounding Decoder.} An ideal scenario for the 3D LMM to perform 3D visual grounding would be to directly output 3D bounding boxes in text format, just like the 2D LMM. However, while the 3D LMM can accurately output the object dimensions represented by a given 3D coordinate, it struggles to directly output the 3D position of the target object. We attempt to train the LLaVA-3D model to directly output the coordinates of the target object in the 3D visual grounding task in text format or using special tokens~\cite{3dllm}, resulting in   7.8 Acc@0.25 and 8.2 Acc@0.25 on the ScanRefer benchmark. However, utilizing our grounding decoder could achieve a performance of 50.1 Acc@0.25.


\vspace{3pt}
\noindent\textbf{Generalization to various 2D LMMs.} In Tab.~\ref{exp:comparison_on_arch}, we adapt our LLaVA-3D to other 2D LMMs, such as LLaVA-1.5~\cite{llava-1.5} (d) and InternVL2.5~\cite{internvl} (e). Experiments demonstrate that our method could be a general extension to equip 2D LMMs with 3D capabilities due to the simple yet effective designs. Comparing (d) and (e), we can observe that LLaVA-3D can benefit from stronger 2D understanding capabilities of the 2D LMM base model. Besides, due to the intrinsic consistency between multi-view image 3D scene representation and video, the results in (e, f) demonstrate that our method can enjoy the significant benefits from video LMMs.

\section{Conclusion}

We propose LLaVA-3D, a simple yet effective framework built upon the well-established LLaVA. LLaVA-3D extends LLaVA's capabilities to perform 3D scene understanding and grounding by leveraging the 3D patches and grounding decoder while efficiently preserving the 2D visual understanding and reasoning capability. Experimental results show that our method achieves state-of-the-art performance on various 3D tasks and benchmarks. We hope that our model will inspire new ideas for building 3D LMMs, and we plan to explore the application of LLaVA-3D in more downstream scenarios, such as robot manipulation and navigation.

{
    \small
    \bibliographystyle{ieeenat_fullname}
    \bibliography{main}
}

\clearpage
\setcounter{page}{1}
\appendix
\maketitlesupplementary

\section{Implementation Details}
\label{sec:exp_settings}

LLaVA-3D is built upon the LLaVA-Video-7B~\cite{llava-video}, utilizing their pre-trained weights from the HuggingFace library, and follows a two-stage training process. Each subsequent stage builds upon the weights learned in the previous stage. The number of views $V$ is set to 32. When adapting our method to LLaVA-1.5~\cite{llava-1.5}, due to the LLM context length limitation, we use the voxelization pooling to compress the 3D patch token numbers, and the maximum number of 3D patch tokens after 3D pooling is set to 3096. All experiments are conducted on 16 × 80G A100 GPUs. 

\vspace{4pt}
\noindent\textbf{Settings of Stage 1.} We use the Adam optimizer to train our model for one epoch with a total batch size of 16 and a warmup ratio of 0.03. During the warmup phase, the learning rates peak at 1e-5 for the LLM, 3D position encoding layer and grounding decoder, and 2e-6 for the vision encoder. The training objectives consist of the auto-regressive language modeling loss and the grounding decoder training loss.

\vspace{4pt}
\noindent\textbf{Settings of Stage 2.} In stage 2, we freeze all the components except for the grounding decoder. The model undergoes 40 training epochs on 16 A100 GPUs with a peak learning rate of 1e-4.

\section{Training Convergence Speed}
\begin{figure}[t]
  \centering
   \includegraphics[width=0.9\linewidth]{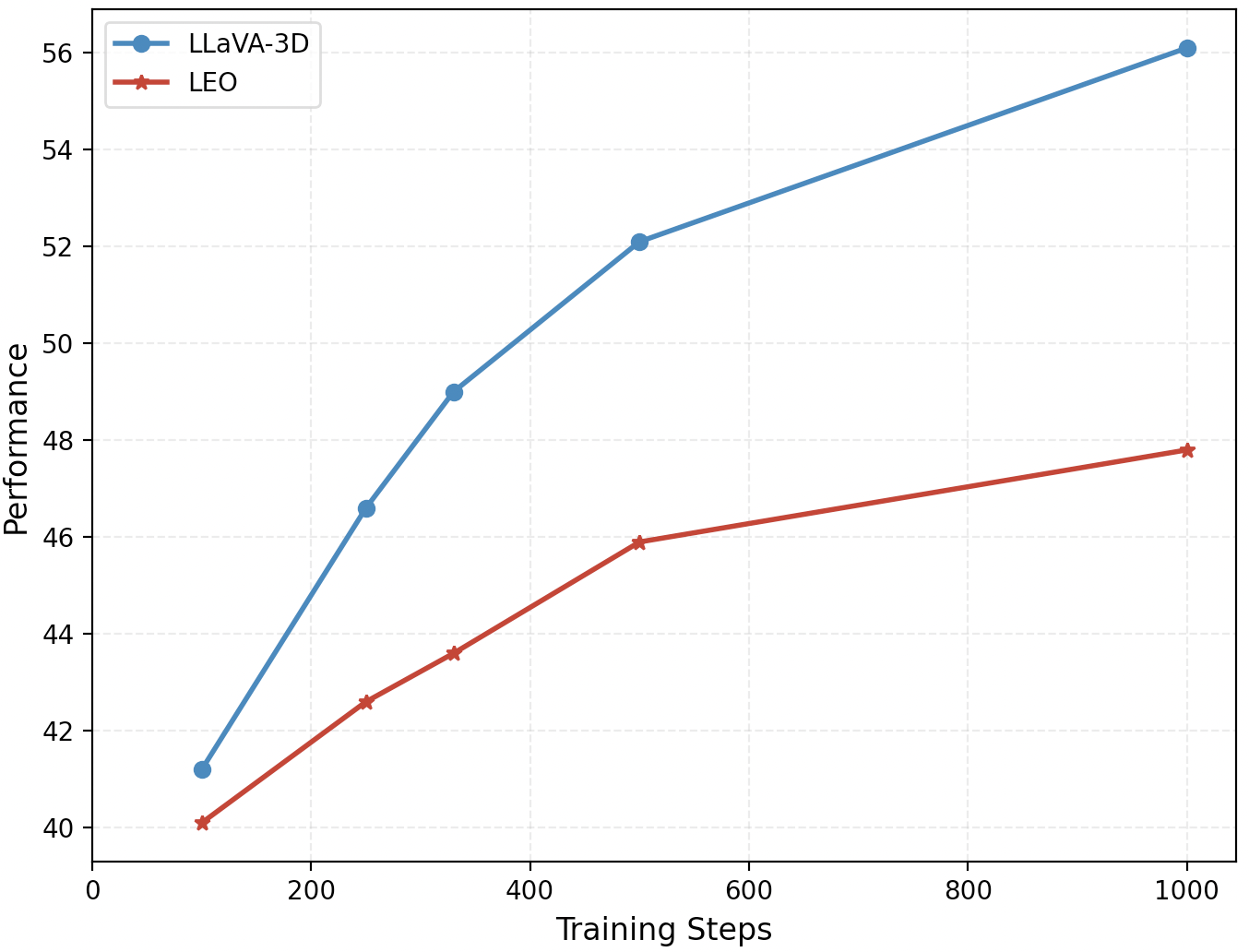}
   \captionsetup{aboveskip=10pt}\captionsetup{belowskip=0pt}\caption{\textbf{Training convergence comparison.} LLaVA-3D achieves higher data efficiency and faster convergence speed during the instruction tuning stage compared with existing 3D LMM: LEO.} 
   \label{fig:convergence}
\end{figure}

To further validate the effectiveness of 2D LMM-based Architecture and ensure fairness as much as possible, we choose LLaVA-1.5 as the base model and replace the LLaVA-3D-Instruct-86K dataset in stage 1 with the MMScan QA~\cite{mmscan} training data. We record and evaluate the performance of LLaVA-3D under different training data ratios. Besides, we further fine-tune LEO~\cite{leo} on full MMScan QA training data based on the officially released model checkpoint. Both models utilize Vicuna-7B as the LLM, ensuring comparable parameter counts. As illustrated in Fig.~\ref{fig:convergence}, LLaVA-3D surpasses LEO's full-step performance even when trained on less than 300 steps, indicating better data efficiency and 3.5x faster training convergence speed.

\section{More Architecture Details}


In this section, we provide more details about how to connect LLM with the grounding decoder via special token embeddings, which enables the end-to-end optimization of the entire model and the grounding decoder details.

\subsection{Connecting with Grounding Decoder} For 3D understanding tasks like 3D visual grounding, we employ the grounding decoder to localize objects according to the user query. Specifically, we introduce a special localization token \texttt{<LOC>} into the LLM vocabulary, and the LLM is trained to predict the special location token to represent the 3D bounding boxes prediction when the task necessitates 3D bounding box outputs. The last layer embedding of this location token is then sent to the 3D grounding decoder as a condition. Grounding Decoder receives both the 3D patch features and the obtained location token embeddings as inputs and predicts the 3D visual grounding results.

\subsection{Grounding Decoder Details} 
\begin{figure}[t]
  \centering
   \includegraphics[width=0.7\linewidth]{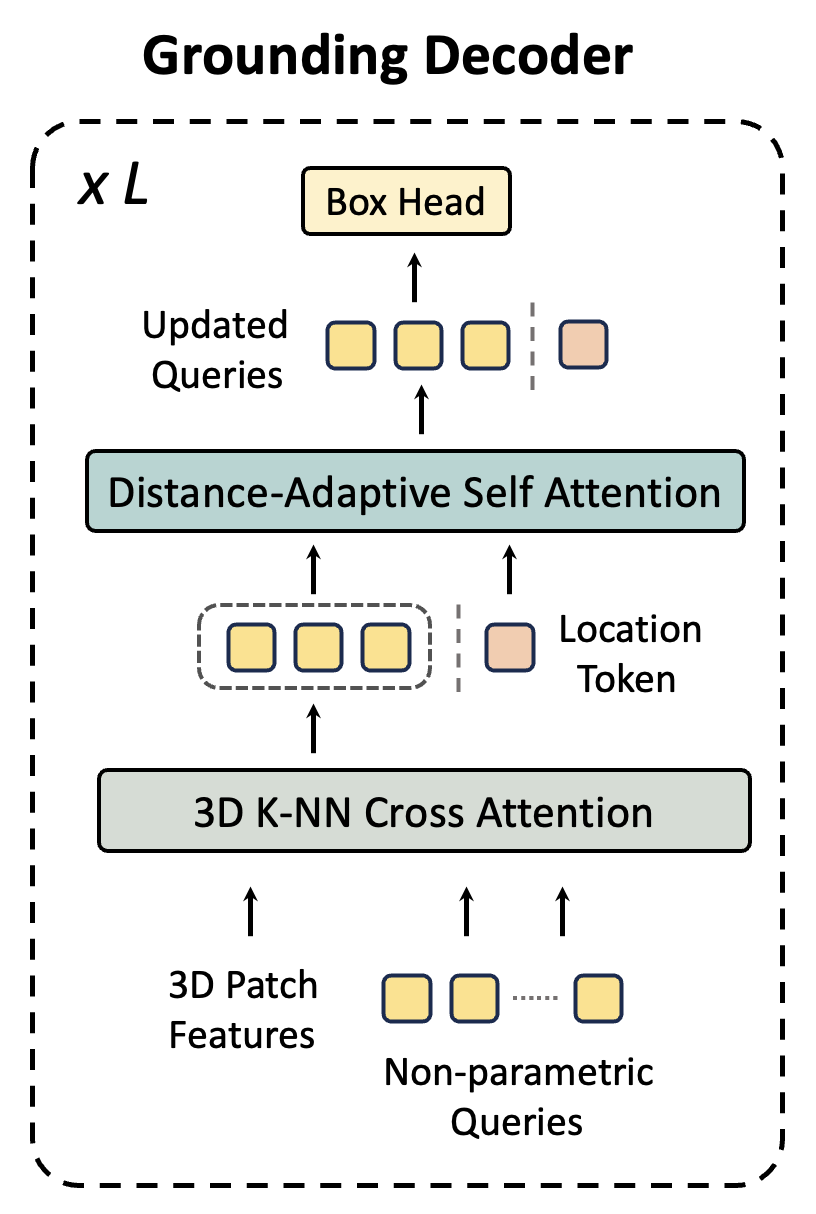}
   \captionsetup{aboveskip=10pt}\captionsetup{belowskip=0pt}\caption{\textbf{Grounding Decoder Architecture.}} 
   \label{fig:grounding_decoder}
\end{figure}

Here illustrate more architectural and training objective details about the grounding decoder. Our grounding decoder consists of $L=4$ decoder layers, as illustrated in Fig.~\ref{fig:grounding_decoder}. For query initialization, we employ farthest point sampling to select $N=512$ instance queries from the 3D patches. We initialize the value of queries to zeros and only the 3D coordinates of the sampled queries are used to set the corresponding learnable positional encoding. 

\paragraph{Multi-Scale 3D k-NN Cross Attention.} During the cross-attention between the queries and 3D patch features, the instance queries can only attend to the features of $k$ nearest 3D patch neighbors to accelerate training convergence speed and reduce the memory usage. To capture the object geometry information at different scales, the $k$ is set to be $\{16, 32, 64, 128\}$ in different decoder layers. Inspired by ODIN~\cite{odin}, we encode the relative position between the query and its neighbor into the position encoding with a learnable MLP. The position encoding is incorporated into the cross-attention computation by adding it to the instance queries and their corresponding 3D patch neighbors.

\paragraph{Distance-Adaptive Self Attention.} After attending to the 3D patch features, we utilize the distance-adaptive self-attention layer~\cite{3d-llava} to model the relative spatial relationship among the queries and achieve the visual-language feature communication. Distance-adaptive self-attention introduces a bias term based on the distances between instance queries. The pairwise attention between the i-th instance query and the j-th instance query is computed as:

\begin{equation}
    Attn(Q_i, K_j, V_j) = Softmax(\frac{Q_iK_j^T}{\sqrt{C}} - \sigma \cdot D)V_j,
    \label{eq:1}
\end{equation}

where $Q, K, V$ is the query, key, and value of the attention module, $C$ is the channel of the embedding, $\sigma$ is a learnable parameter based on the query, and $D$ indicates the Euler distance between the position of these two instance queries. For the location token that does not have the 3D coordinate information, the bias term between this location token and the instance queries is set to zero.

\paragraph{Box Head.}
The box head consists of a two-layer MLP, which takes the updated instance queries in each decoder layer as input and predicts the corresponding 3D bounding boxes. 

\subsection{Training Objective}

After matching the instance queries with ground truth 3D bounding boxes, for each match between a proposal and a ground truth object, we compute the DIOU loss~\cite{unit3d} between predicted and ground truth boxes. We utilize InfoNCE loss~\cite{infonce} to optimize the similarity between the matched queries and the location token.

\section{More Components Analysis}
\label{sec:ablation}

To better understand the impact of different components and the generalizability of our LLaVA-3D, we conduct a thorough ablation study on the ScanQA and SQA3D benchmarks based on LLaVA-1.5~\cite{llava-1.5}.

\vspace{3pt}
\noindent\textbf{Impact of Pooling Strategy.} 
Here we conduct various experiments to evaluate the effects of the different pooling strategies. For voxelization pooling, we adopt the simple voxelization approach from ODIN~\cite{odin}. As shown in Tab.~\ref{tab:pooling_strategy}, the voxelization pooling strategy outperforms the FPS pooling method on 3D QA benchmarks. Model performance can be improved by either decreasing voxel size in voxelization pooling or increasing the number of 3D patch tokens in FPS pooling.

\begin{table}
    \caption{
        \textbf{Comparsion on different pooling strategies}.}
    \centering
    \resizebox{\linewidth}{!}{
        \begin{tabular}{ccccc}
        \toprule
        Pooling Strategy  & Voxel Size & Token Number & ScanQA   & SQA3D \\ \hline
        Voxelization     & 0.4 & Dynamic        & 24.1 & 53.2                 \\
        Voxelization     & 0.3 & Dynamic        & 25.9 & 54.8               \\
        \rowcolor{gray!20} Voxelization     & 0.2 & Dynamic        & 27.0 & 55.6             \\
        FPS    & -  & 576 & 25.7          & 54.9  \\ 
        FPS     & -  & 1024 & 26.3         & 55.2 \\ 
        \bottomrule
        \end{tabular}
    }
\label{tab:pooling_strategy}
\end{table}

\vspace{3pt}
\noindent\textbf{Multi-View Images Sampling Strategy.} 
To balance computational efficiency with visual coverage, we sample $V$ views from the egocentric images of each 3D scene. We investigate two sampling strategies during inference: \textit{Uniform Sampling}, which evenly samples images across the scene, and \textit{Text-Guided Sampling}, which selects frames based on CLIP image-text similarity scores to the input instruction. Since our experiments show a similar performance, we adopt uniform sampling for its simplicity.

\vspace{3pt}
\noindent\textbf{Number of Views.} An intuitive assumption is that sampling more views from the 3D scene will preserve more information about the 3D scene. We conduct a comparative experiment varying the number of views sampled from 3D scenes. Tab.~\ref{tab:sample_num} presents the Exact Match (EM) scores on ScanQA and SQA3D across different settings, revealing that the increase in EM score is marginal as the number of views increases. Additionally, the experimental results indicate that exceeding a certain number of views can degrade the model's performance.

\begin{table}
    \caption{
        \textbf{Comparison on performance on 3D QA tasks under different number of multi-view images}.}
    \label{tab:ablation-click-scanqa}
    \centering
    \resizebox{0.95\linewidth}{!}{
        \begin{tabular}{cccc}
        \toprule
        Number of Views  & Number of Tokens & ScanQA   & SQA3D \\ \hline
        16     & 9216         & 26.2 & 55.1              \\
        \rowcolor{gray!20} 20     & 11520         & 27.0 & 55.6             \\
        24     & 13824       & 27.0 & 55.4                 \\
        40     & 23040   &26.7 &55.2  \\ 
        \bottomrule
        \end{tabular}
    }
\label{tab:sample_num}
\end{table}

\section{More Qualitative Results}



\paragraph{3D Scene Understanding.}

\begin{figure*}[!htbp]
  \centering
   \includegraphics[width=0.9\linewidth]{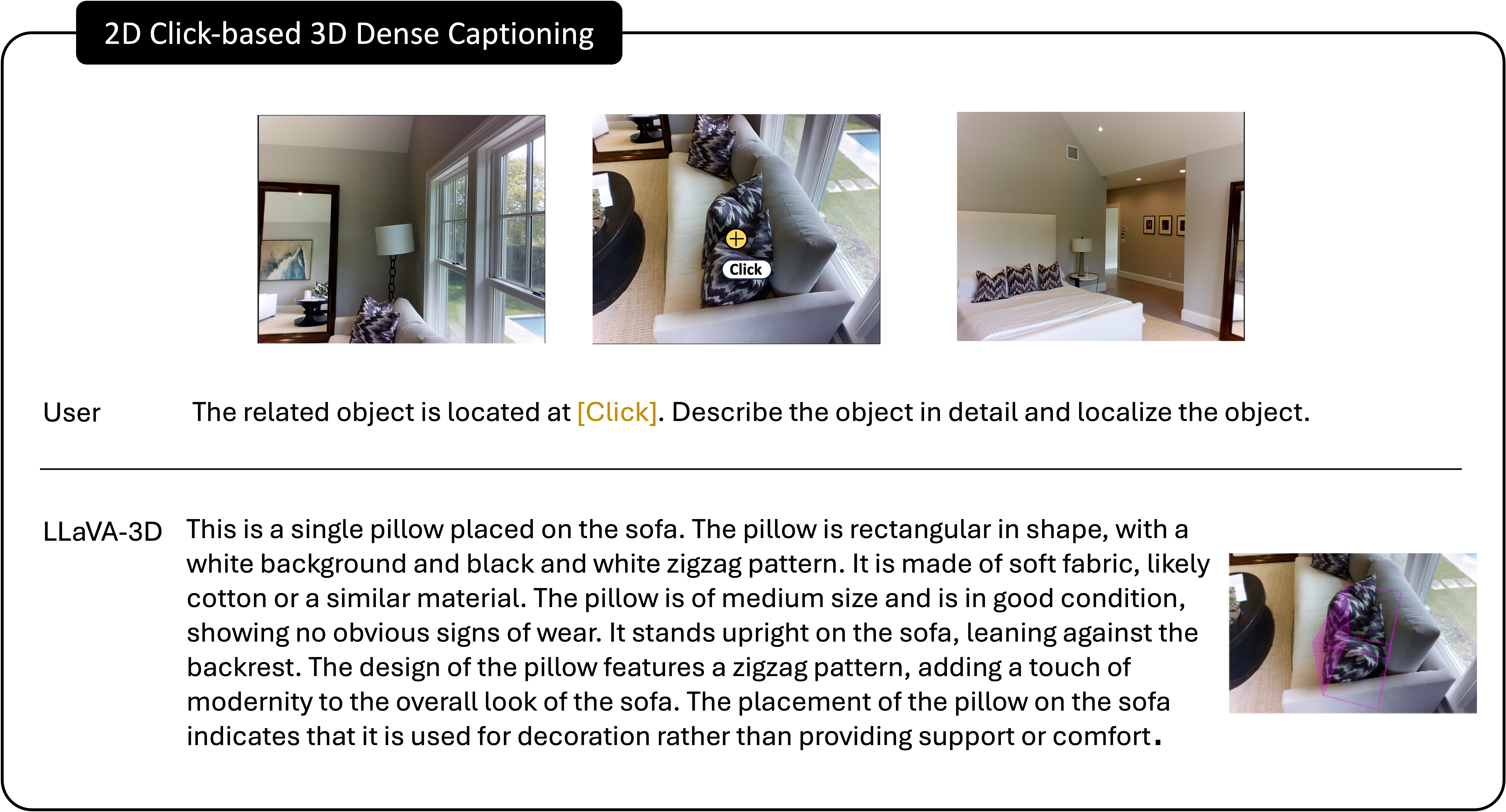}
   \captionsetup{aboveskip=5pt}\captionsetup{belowskip=0pt}\caption{LLaVA-3D could perform 2D Click-based 3D dense captioning, generating the corresponding object caption and 3D bounding box.} 
   \label{fig:demo_1}
\end{figure*}

\begin{figure*}[!htbp]
  \centering
   \includegraphics[width=1\linewidth]{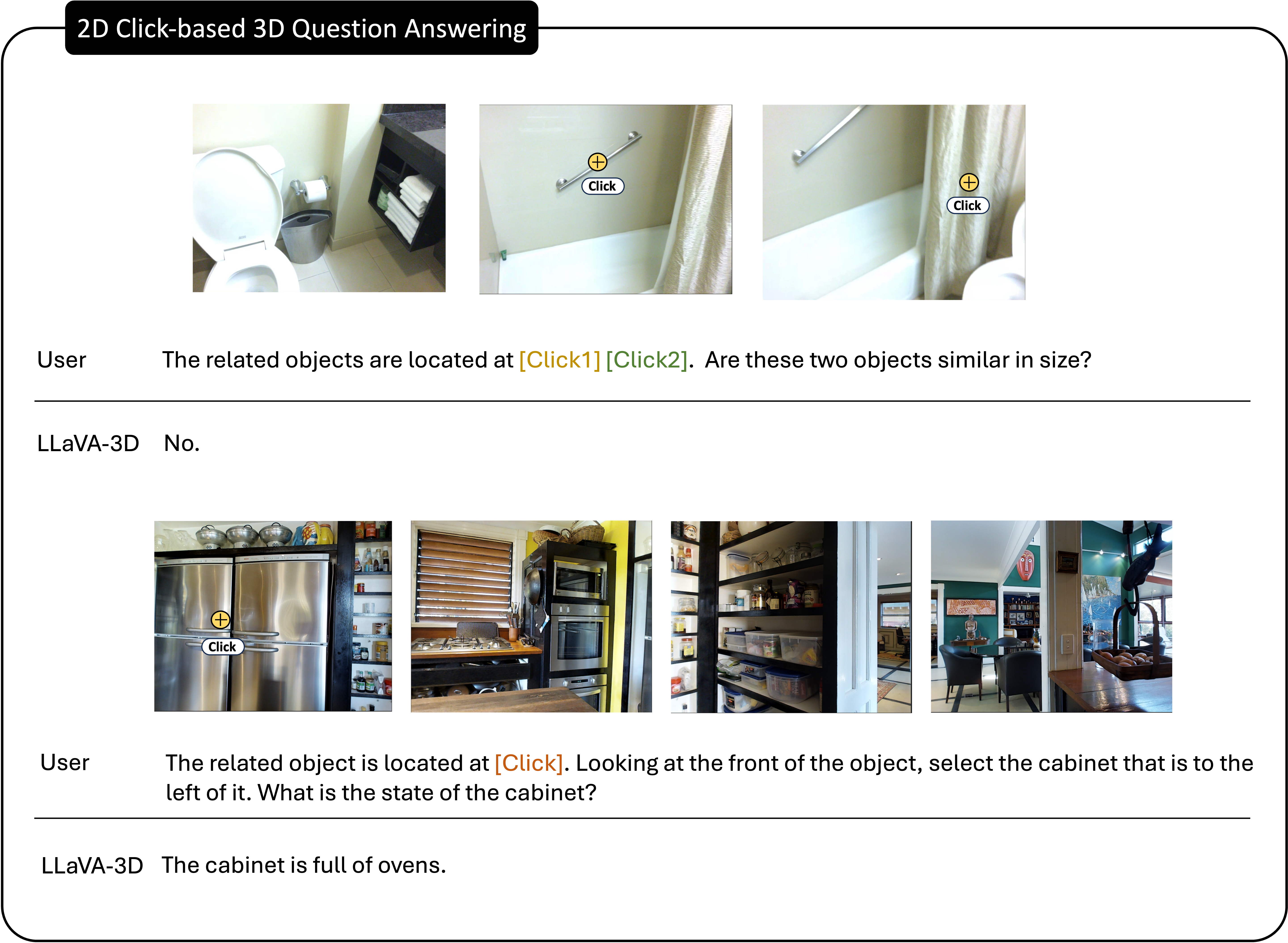}
   \captionsetup{aboveskip=5pt}\captionsetup{belowskip=0pt}\caption{LLaVA-3D could perform 2D Click-based 3D question answering, now users could click on the 2D images and ask the question.} 
   \label{fig:demo_2}
\end{figure*}

\begin{figure*}[!htbp]
  \centering
   \includegraphics[width=1\linewidth]{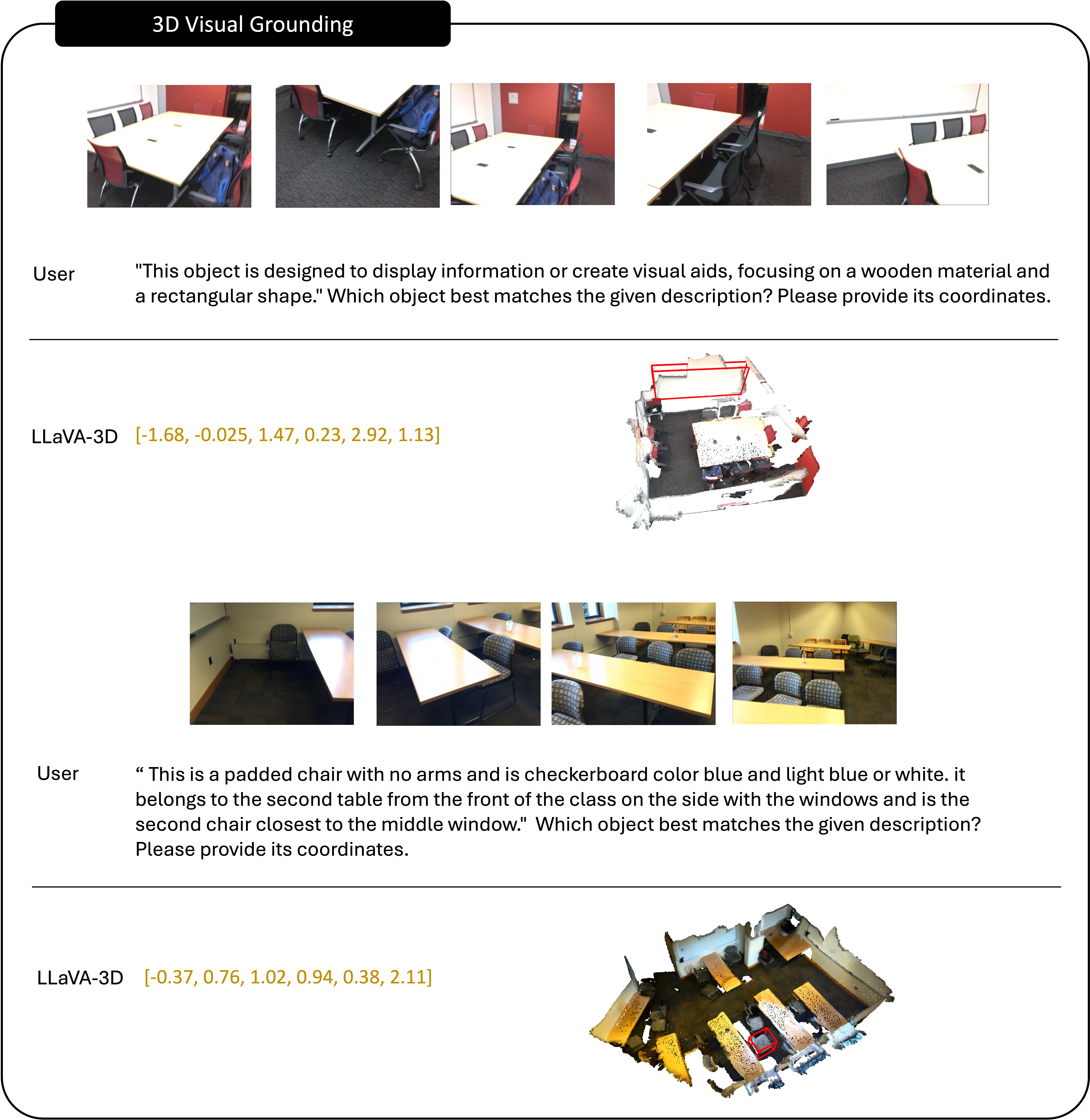}
   \captionsetup{aboveskip=5pt}\captionsetup{belowskip=0pt}\caption{LLaVA-3D exhibits powerful 3D visual grounding capability, enabling accurate 3D bounding boxes output.} 
   \label{fig:demo_3}
\end{figure*}

We evaluate LLaVA-3D on various 3D scene understanding tasks and display
more visualization results from Fig.~\ref{fig:demo_1} to Fig.~\ref{fig:demo_3}. These examples demonstrate LLaVA-3D's robust 3D understanding abilities: comprehensive 3D scene understanding, accurate object recognition, and precise object localization in the 3D world. Besides, our model enables the users to more easily interact with the 3D scene through the 2D images.

\section{Video Demo Comparision}

\begin{figure*}[!htbp]
  \centering
   \includegraphics[width=1\linewidth]{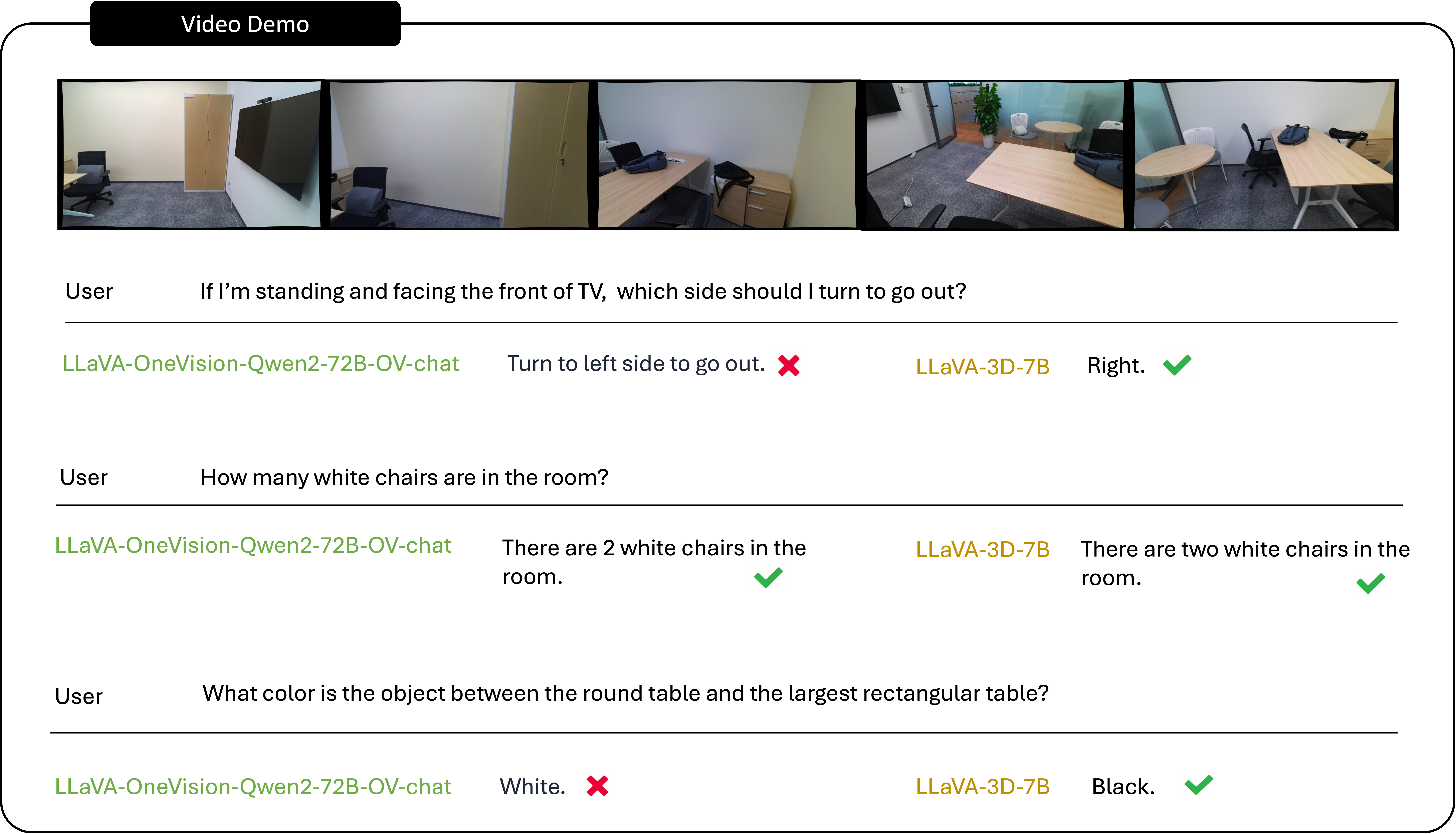}
   \captionsetup{aboveskip=5pt}\captionsetup{belowskip=0pt}\caption{LLaVA-3D achieves superior performance in 3D spatial reasoning and relationship understanding between objects with significantly fewer parameters compared with powerful LLaVA-OneVision 72B.} 
   \label{fig:video_demo}
\end{figure*}

To enhance real-world applicability, we design our framework to process 2D videos - a widely accessible data format that users can capture with standard mobile devices. Our pipeline processes these inputs by uniformly sampling 32 frames and leveraging DUST3R~\cite{dust3r}, an efficient offline MVS method, to obtain depth maps, camera parameters, and poses. Notably, DUST3R completes this process within one minute, enabling seamless conversion of conventional video inputs into our model's required format.
To validate our approach, we conduct comprehensive evaluations against LLaVA-OneVision 72B~\cite{llava-onevision} in Fig.~\ref{fig:video_demo}, a state-of-the-art multimodal model that demonstrates strong capabilities across diverse 2D scenarios, including single-image understanding, multi-view reasoning, and video understanding. The qualitative results reveal that our method achieves superior performance in 3D spatial reasoning and relationship understanding between objects with significantly fewer parameters (7B), highlighting the effectiveness of our 3D-aware architecture.




\end{document}